\documentclass[final,3p]{elsarticle}

\usepackage{lineno,hyperref}
\modulolinenumbers[5]

\usepackage{lineno}
\usepackage{graphicx}
\usepackage{algorithm}
\usepackage{algorithmic}
\usepackage{enumerate}
\usepackage{rotating}
\usepackage{color, colortbl}
\usepackage{array}
\usepackage{mdwmath}
\usepackage{mdwtab}
\usepackage{eqparbox}
\usepackage{verbatim}
\usepackage[tight,footnotesize]{subfigure}
\usepackage{multirow}
\usepackage{url}
\usepackage{amsfonts}%
\usepackage{stfloats}
\usepackage{balance}
\usepackage{flushend}
\usepackage[table]{xcolor}
\usepackage{colortbl}
\usepackage{graphicx}
\usepackage{caption}
\usepackage{mathtools}
\usepackage{subfigure}
\usepackage{stfloats}
\usepackage{enumitem}

\journal{Vehicular Communications}

\biboptions{numbers,sort}

\begin{document}

\begin{frontmatter}

\title{A Swarm Algorithm for Collaborative Traffic in Vehicular Networks}

\author{Jamal Toutouh\corref{mycorrespondingauthor}}
\cortext[mycorrespondingauthor]{Corresponding author}
\ead{jamal@lcc.uma.es}
\author{Enrique Alba}
\address{Dept. de Lenguajes y Ciencias de la Computaci\'on, University of Malaga, Malaga, Spain}

\begin{abstract}
Vehicular ad hoc networks (VANETs) allow vehicles to exchange warning messages with each other.
These specific kinds of networks help reduce  hazardous traffic situations and improve safety, which are two of the main objectives in developing Intelligent Transportation Systems (ITS). For this, 
the performance of VANETs should guarantee the delivery of messages in a required time.
An obstacle to this is that the data traffic generated may cause
network congestion.
Data congestion control is used to enhance network capabilities, increasing the reliability of the VANET by decreasing packet losses and communication delays.
In this study, we propose a swarm intelligence based distributed congestion control strategy to maintain the channel usage level under the threshold of network malfunction, while keeping the quality-of-service of the VANET high.
An exhaustive experimentation shows that the proposed strategy improves the throughput of the network, the channel usage, and the stability of the communications in comparison with other competing congestion control strategies.\end{abstract}


\end{frontmatter}


\section{Introduction}
\label{sec:introduction}
\vspace{-0.1cm}
Over the last few decades, the synergistic utilization of \emph{information and communication technologies} (ICT) in vehicular environments has revolutionized the automotive industry. 
This has encouraged the emergence of a great variety of new services based on \emph{Intelligent Transportation Systems} (ITS) focused on improving road safety and travelers' experience.  
%
Most of these great advances rely on vehicular networks that allow the periodic exchange of messages between the different agents that are part of road transportation (e.g., vehicles or elements of the infrastructure)~\cite{VAHDATNEJAD201643}.
This communication technology is commonly known as \emph{vehicular ad hoc networks} (VANETs), which are principally composed by vehicles equipped with wireless interfaces in their on-board units (OBUs) that allow direct short range communications (DSRC) by utilizing the wireless access in vehicular environments (WAVE) standards, i.e., IEEE 802.11p and IEEE 1609~\cite{Campolo2015}.

VANETs are applied to deploy ITS to provide a large number of smart mobility services and applications.
The most important category of applications based on VANETs are designed to provide safe environments for road travel and intelligent road traffic management. Those are are known as \emph{cooperative vehicle safety} (CVS) and \emph{traffic efficiency applications}, respectively~\cite{DIAS201422,CHAQFEH2014214}.

\begin{figure}[!h]
\setlength{\abovecaptionskip}{0.0pt}
\setlength{\belowcaptionskip}{0.0pt}
\begin{center}
\vspace{-0.1cm}
\includegraphics[width=0.5\linewidth]{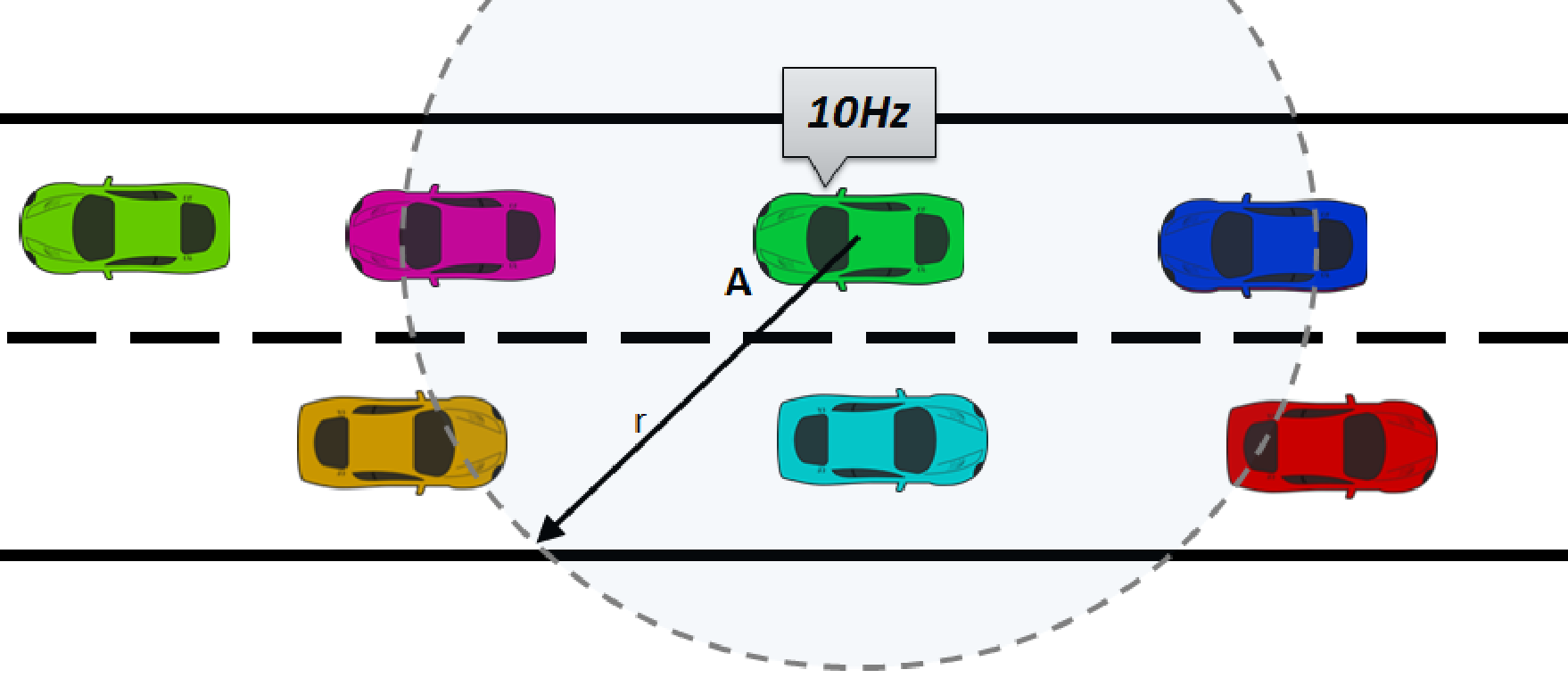}
\vspace{-0.1cm}
\end{center}
\caption{\emph{Car A} performs CVS communication.}
\vspace{-0.1cm}
\label{fig:ejemplo-basico}
\end{figure}

CVS principally relies on exchanging short messages (known as \emph{beacons}) through the DSRC channel~\cite{Fallah2010}.
These messages are broadcasted in the neighborhood (1-hop) defined by the communication range of the nodes ($r$).
Beacons include vehicle kinematics and other relevant information.
VANET nodes are continuously broadcasting beacons (\emph{beaconing}) with a given \emph{beacon frequency} or \emph{beacon rate} (see Figure~\ref{fig:ejemplo-basico}).

A challenging issue in the deployment of CVS is the network congestion when the scale of the system grows.
This is mainly due to the critical increase of the periodic beacons, which generates a heavy communication load. 
Congestion increases packet losses and communication delays, i.e., it degrades the performance and the quality-of-service (QoS) of the VANETs. 
This may lead to excessive information inaccuracy and eventually failure of CVS \cite{DIAS201422,GUPTA2015223}.

There are many techniques for improving congestion control in VANETs~\cite{Sattari2012}.
One promising line of research is to adapt the broadcasting protocol to the current channel resources by changing its configuration parameters.
Most of studies propose some management of the transmission power (communication range) and the beacon frequency.

In this article, a swarm intelligence based congestion control method (Swarm FREDY) is defined. 
This novel method is stochastic, dynamic, and fully distributed. 
When the VANET uses Swarm FREDY, each node runs its own instance of the algorithm, as a particle of the swarm, to efficiently adapt its beacon rate to the available channel capacity and cooperate with the neighbor nodes in their congestion control operation.

Swarm intelligence comprises a set of nature inspired artificial intelligence methods in which a set of simple agents mimic the behavior of social organisms (ant colonies, bird flocking, fish schooling, etc.) in a given algorithm~\cite{yang2013swarm}. 
This kind of methods has been successfully applied in many hard to solve optimization problems, such as 
robotics~\cite{Brambilla2013},
engineering~\cite{rahman2015swarm},
telecommunications~\cite{GIAGKOS201423}, and  
machine learning~\cite{Salama2015}.

%
The research questions our study addresses are:
\begin{itemize}[noitemsep]
\setlength\itemsep{0em}
\item[\textbf{RQ1}] \textit{What is the true importance of controlling the communication congestion for road safety?}
\item[\textbf{RQ2}] \textit{Can a lightweight swarm based algorithm of constant complexity provide competitive congestion control in vehicular communications?}
\item[\textbf{RQ3}] \textit{Can a stochastic method be more competitive than the existing deterministic ones?} 
\end{itemize}

In short, the main contributions of this study are:
$i)$ defining the optimization problem of congestion control by beacon frequency adaptation
and $ii)$ proposing a swarm based congestion control method.

The rest of this article is organized as follows.
Section~\ref{sec:related-work} summarizes the state of the art in the field of congestion control in vehicular networks.
Section~\ref{sec:fbr} introduces the concept of fair beaconing in VANETs.
Section~\ref{sec:optimization-problem-definition} defines the fair beacon rate optimization problem. 
Section~\ref{sec:fredy-definition} presents the Swarm FREDY method proposed here.
Section~\ref{sec:evaluation} describes the experimental evaluation framework and Section~\ref{sec:experimental-results} analyzes the numerical results.
Finally, Section~\ref{sec:conclusions} outlines our conclusions and the main lines of the future research.

\vspace{-.1cm}
\section{Related Work}
\label{sec:related-work}
 \vspace{-.1cm}

Congestion control is an important research topic with the objective of providing reliable environments in modern network communications~\cite{Lochert2007}.
If we focus on vehicular communications, congestion control is an even more critical concern~\cite{CHAQFEH2014214}.
The reliability of CVS,
which could make the difference between saving lives or not,
is highly dependent on two quality-of-service (QoS) metrics: the packet loss and the communication delays.
Congestion, which occurs when the network load exceeds the capacity of the network links,
generally leads to a deterioration of these two metrics. 

Several strategies have been proposed to address congestion problems in VANETs, 
keeping the communication capabilities of the nodes over a given QoS threshold.
Most of them can be included in the following basic schemes:
$i)$ adapting the transmission range of transmission channels,
$ii)$ adjusting the data rate generation of applications and services,
$iii)$ hybrid methods by combining the two aforementioned schemes,
and $iv)$ scheduling data packets in various channels. 

\sloppy The scheme that includes transmission range adaptation follows the idea that reducing the transmission power of beacons
keeps the network load below a certain threshold for an optimal VANET operation.
However, an excessive reduction of transmission power could cause node isolations when the network density decreases~\cite{Artimy2005,Mittag2008}.

In order to avoid this, the distributed fair power adjustment for VANETs (D-FPAV) dynamically controls the transmission power (range) to keep the beaconing traffic under a threshold called \textit{MaxBeaconingLoad} (MBL)~\cite{Torrent2009}.
As the density of the neighborhood increases, the power transmission is reduced to keep the network load under the MBL.
However, it cannot manage situations in which the MBL is violated and the transmission is already at the minimum allowed power.

Controlling congestion by adjusting the beacon rate (the second basic scheme), is similar to the adaptation of the transmission range.
The idea is to find the optimal beaconing rate for each node that suits the applications and the network status.
On the one hand, a high beacon rate increases the accuracy of the system's knowledge, which is important for safety services. However, congestion is more likely to appear in these cases.
On the other hand, a low beaconing rate increases the latency of the information,
but prevents VANETs from overloading the communication channel. 
Therefore there are trade-offs throughout.

In an initial approach, the rate of beacon generation was adjusted according to the information about the current speed of the node, the failure of the transmission attempt, and the beacon reception success rate~\cite{Xu2004}.
A lookup table stores the predefined beaconing rates in terms of the three metrics analyzed.
The rate decreases whenever
the vehicles reduce their speed,
the maximum failed transmission attempts is observed or
the minimum reception success rate is measured.
The main drawback of this method is that VANETs are fully distributed systems by nature, and it is not clear how to obtain the necessary statistics (such as the the reception rate).

Later, an adaptive beaconing communication scheme for cooperative active safety system (CASS) was introduced~\cite{Rezaei2007}. 
In this method, the nodes broadcast kinematics information to all neighbors within a certain communication range. 
Each node uses a so-called \textit{self estimator} to estimate its own position, speed, and heading.
In turn, it runs a \textit{remote estimator} to mimic the information about its own position from the perspective of neighboring vehicles.
The idea is to broadcast beacons only when the difference between the calculations of both estimators exceeds a maximum deviation threshold. 

In \cite{Schmidt2010}, the authors studied the \textit{situation-adaptive beaconing},
a rate control method based on the movement of the own node and neighboring vehicles, including microscopic aspects (e.g., speeds) and macroscopic aspects (e.g., road traffic density).
They analyzed different adaptation schemes based on: vehicle's own movement, surrounding vehicles' movement, and a combination of the previous ones.
They concluded that the aggregation of several schemes provides a better congestion control than using them individually. 

Taking into account the third congestion control scheme, some researchers defined hybrid congestion control methods that apply transmission power and rate adaptations together to overcome the limitations of both mechanisms.
These methods first address the congestion control of the medium by utilizing beaconing rate adaptation.
If the current beacon rate generation is under a minimum threshold for the actual requirements and still the congestion situation is not avoided,
then a power transmission adjustment is also applied~\cite{Djahel2012,Tielert2013}.

Some studies have introduced the use of metaheuristics to define hybrid congestion control methods.
They use these techniques to find efficient parameterizations,  
after a congestion is detected.
Along these lines, 
the Single-Objective Tabu Search was proposed to minimize the communication delay only~\cite{Taherkhani2012}.
Afterwards, a Multi-Objective Tabu Search was applied to minimize the communication delay and jitter~\cite{Taherkhani2015}.
Their main issue was the relatively high computation complexity (run times),
that needed to be considered as part of the final communication delays.

Finally, some authors have proposed scheduling beacons through several channels depending on their current availability. 
In~\cite{Zeeshan2015}, the authors proposed the QoS-aware radio access technology (RAT), a specific congestion control method for heterogeneous vehicular networks (HVN), i.e., nodes are equipped with WAVE and LTE cellular network interfaces.
The QoS-aware RAT applies an iterative method to keep the network load under a given threshold to avoid network congestion.
For this, when the network load grows, RAT reduces the beaconing rate until reaching a given threshold.
If the current rate is under a given QoS threshold (it cannot be reduced anymore),
the beacons are broadcast via LTE. 

All these proposals have different drawbacks that prevent their use in real VANETs: 
they rely on information that is not available in the current standards,
they require the use of a central entity (while VANETs are fully distributed),
their computational cost prevents their application because of a critical increment in the communication delays,
and so forth.
In the present study, we propose different stochastic, dynamic, and distributed congestion control strategies to
increase the communication reliability and balance in VANETs, while avoiding the previous shortcomings.
The main goal is to efficiently manage the broadcasting of beacons because they generate the predominant (almost only) overheads in the control channel.
This work is an extension of our preliminary study presented in~\cite{Toutouh2016}, in which we presented the Distributed Intelligent Fair Rate Adaptation (DIFRA) algorithm family.
DIFRA is a set of basic greedy methods that adapt the beaconing frequency of each vehicle according to the current channel load. Let us put them to work in this article.

\section{Fair Beacon Rate Broadcasting in VANETs}
\label{sec:fbr}

This section presents the \emph{fair beacon rate (FBR)} strategy to address the congestion control problem in VANETs whilst also avoiding any starvation of the nodes~\cite{Toutouh2016}.
The main idea is to use the beacon rate as a QoS metric that should be numerically balanced among all the nodes in the neighborhood in order to guarantee the proper operation of the VANET~\cite{Fallah2010}.

The section is organized as follows:
First, the frequency of broadcasting beacons as an important QoS metric for CVS applications is discussed.
Then, FBR use case is illustrated. 

\vspace{-0.2cm}
\subsection{Beacon Frequency as a QoS Numerical Metric}
\label{sec:BF-QoS}
\vspace{-0.05cm}

In VANETs, beacons are broadcasted to neighboring vehicles at regular intervals to make them aware of their environment.
%
Therefore, beacons contain kinematic information of vehicles.
VANET nodes broadcast beacons principally to achieve two goals:
i) a fresh knowledge of their surroundings, to prevent unsafe situations
and ii) internal adjustments of the VANET communication protocols.

The reliability of CVS applications is highly dependent on beacon broadcasting. 
Applications manage more accurate information when the nodes are able to exchange messages with higher resolution (beacon rate).
Thus, the beacon frequency can be used as a QoS metric of the system, since the higher the frequency (without generating congestion) the higher the accuracy of the received information~\cite{Fallah2010}.

Due to the channel's capacity limitations, it is crucial that nodes broadcast beacons at a suitable beacon frequency in accordance with the current network status.
On the one hand, it is strongly accepted that a high beacon rate can easily result in channel congestion in regions of high road traffic density,
therefore causing a high reduction of beacon delivery and a critical 
throughput degradation~\cite{Fallah2010}.
On the other hand, larger intervals between consecutive beacons (lower beacon rate) increase the uncertainty of the CVS applications, i.e., nodes might not know the required information about their neighbors for a certain (too long) time.
%
%
%
%
Thus, the beacon rate can be used as a general QoS metric to represent the reliability of a CVS. In turn, this rate affects the throughput of the VANET.
%

\subsection{Use Case of FBR Utilization in VANETs}
\label{sec:fair-beacon-rate}

Congestion control mechanisms may produce \emph{unfair} situations (nearby nodes with similar network conditions transmit with wildly different beacon rates).
In IEEE 802.11 communications, mechanisms based on the RTS/CTS protocol have been defined to mitigate \emph{unfairness}
when Carrier-Sense Multiple Access (CSMA) is used.
However, these solutions cannot be directly applied in VANET broadcasting since they have been principally designed for end-to-end data flows~\cite{Wischhof2005}.

Congestion in VANETs can be addressed by means of \emph{fair} beaconing.
In this case, \emph{fairness}
can be seen as the situation in which those vehicles (VANET nodes) located near to each other are able to broadcast beacons with similar and high beacon rates, while avoiding network congestion.
Thus, no VANET node with data to transmit suffers from starvation. 

\begin{figure}[!h]
\setlength{\abovecaptionskip}{0.0pt}
\setlength{\belowcaptionskip}{-10.0pt}
\vspace{-0.2cm}
\begin{center}
\includegraphics[width=0.5\linewidth]{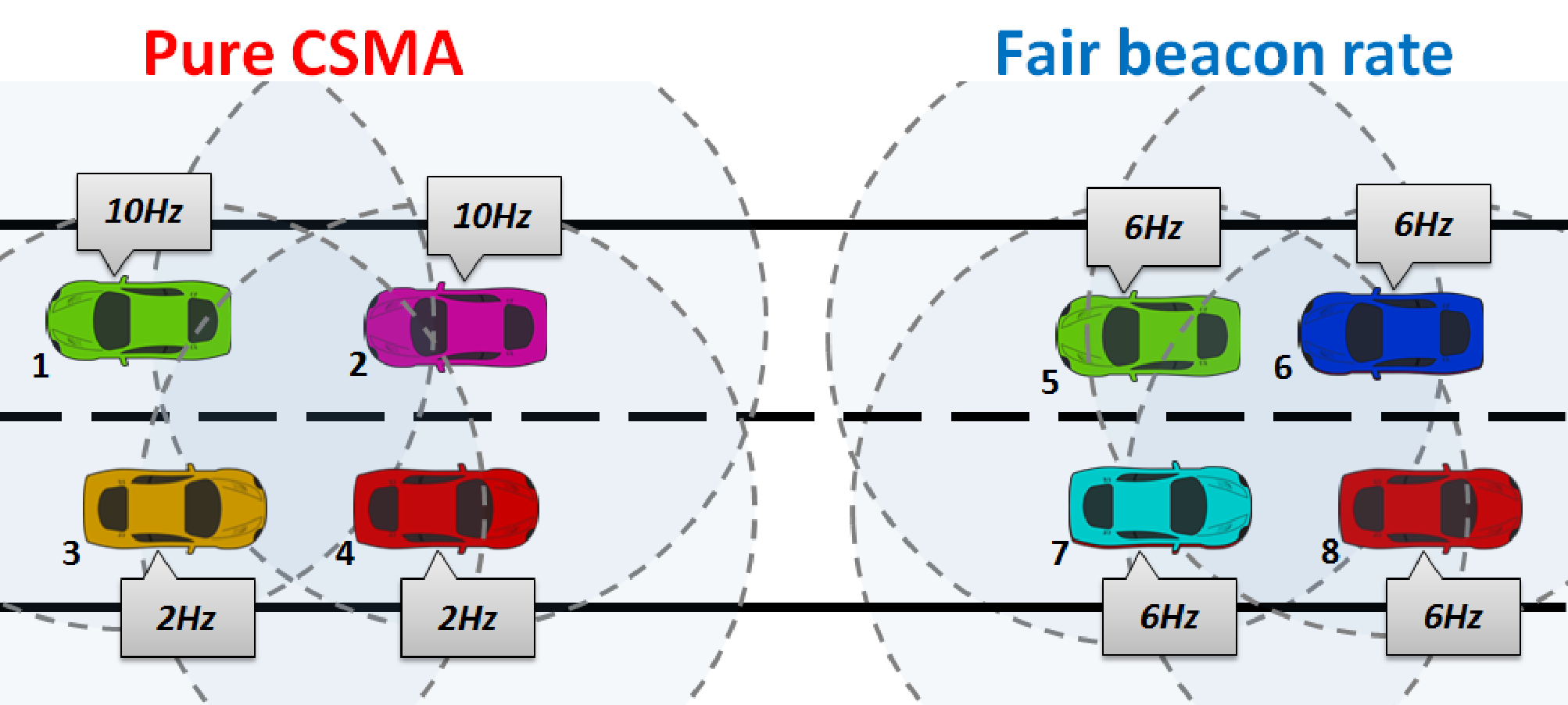}
\vspace{-.0cm}
\end{center}
\caption{Simple deterministic VANET scenario.}
\label{fig:problem-example}

\end{figure}

Figure~\ref{fig:problem-example} shows a very simple CVS example of the difference of using fair beaconing or not.
Here,
it is assumed that
the beacon rates can be adapted from 1 to 10 beacons per second (hertz, $Hz$),
the maximum channel occupancy (or capacity) in terms of beacons per unit time
($MaxQ$) is 30~beacons per second,
the threshold limit ratio over $MaxQ$ that can be used by CVS while still avoiding system malfunction ($\alpha$) is 80\% ($\alpha$=0.8),
and the transmission and the carrier sense have the same range (marked by dotted circles). 
According to $MaxQ$ and $\alpha$ values, the maximum number of beacons per second that can be exchanged through the communication channel is 24 \mbox{($\alpha \cdot MaxQ$=$0.8 \cdot 30$=24)}. This value guarantees the proper operation of the VANET (see Section~\ref{sec:BF-QoS}).

In Figure~\ref{fig:problem-example}, there are two groups of cars located near to each other that represent different situations: \emph{Pure CSMA} and \emph{Fair beacon rate}.
The first one demonstrates the unfair situation resulting from applying a purely CSMA-based method,
which is illustrated by the starvation of cars 3 and 4, which transmit just 2 beacons per second to avoid network congestion.
The main reason is that these two nodes are competing with others that are transmitting beacons at the maximum beacon rate (10~$Hz$). 
This is mainly because cars 1 and 2 are not aware that there are nodes trying to broadcast beacons. 
In the \emph{Fair beacon rate} situation (right-hand side of the figure), all the nodes in the same carrier sense (cars 5, 6, 7, and 8) apply a given mechanism to allow all the nodes to transmit the beacons at the same rate (6~$Hz$) without incurring in congestion. 

In our study, FBR considers three main goals:
$i)$ maintaining the VANET load under a given threshold to avoid network congestion,
$ii)$ avoiding the starvation of nodes that have something to broadcast,
and $iii)$ balancing beacon rates 
(allowing close nodes to exchange beacons with similar rates).

\begin{figure}[!h]
\setlength{\abovecaptionskip}{0.0pt}
\setlength{\belowcaptionskip}{0.0pt}
\vspace{-0.1cm}
\begin{center}
\includegraphics[width=0.4\linewidth]{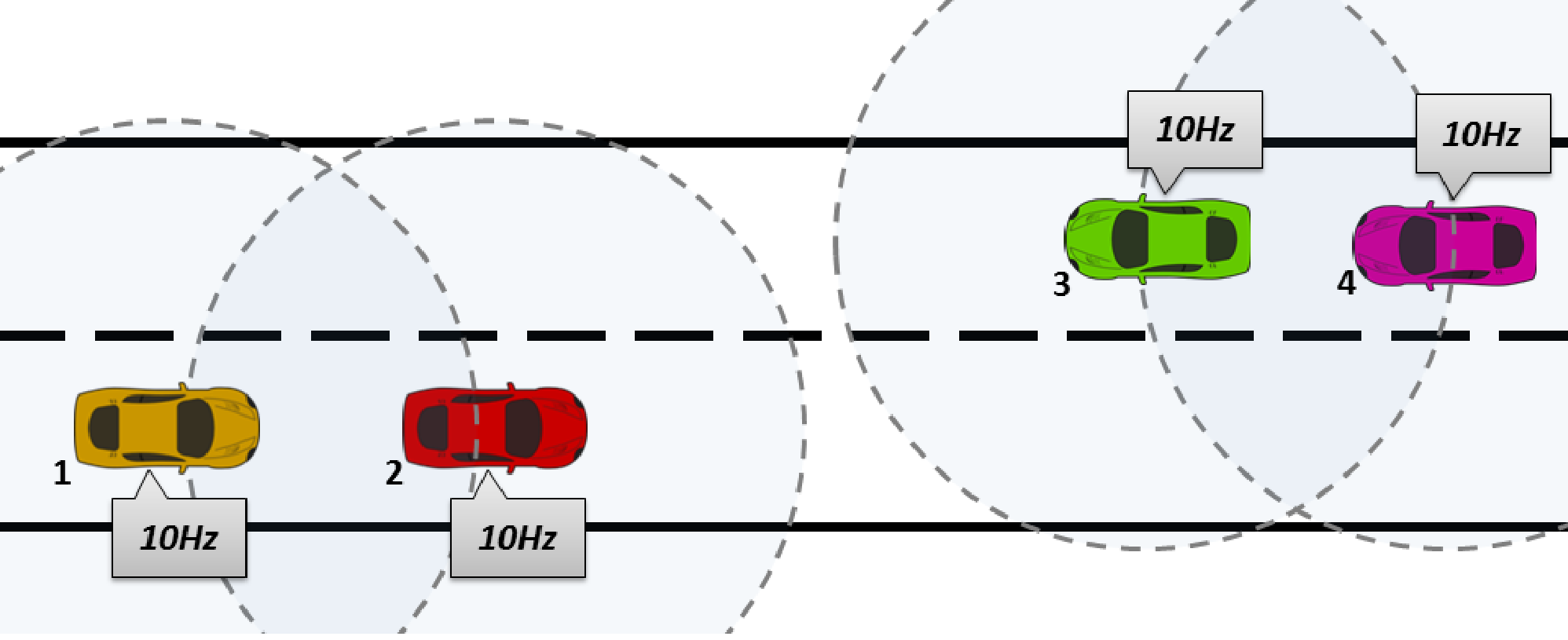}

\small{a) {No beacon rate adaptation is needed.}}
\end{center}

\begin{center}
\includegraphics[width=0.4\linewidth]{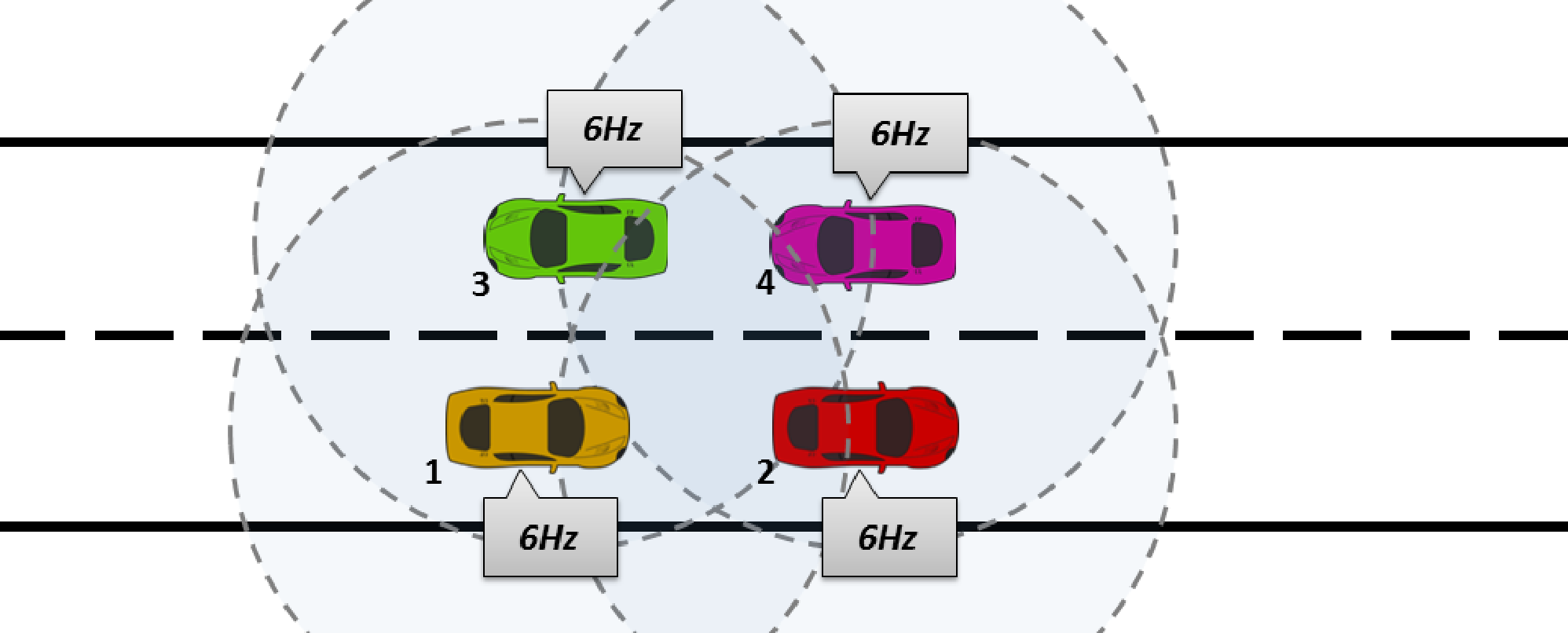}

\small{b) {Beacon rates are adapted to avoid starvation or congestion.}}
\end{center}

\vspace{-0.3cm}
\begin{center}
\includegraphics[width=0.4\linewidth]{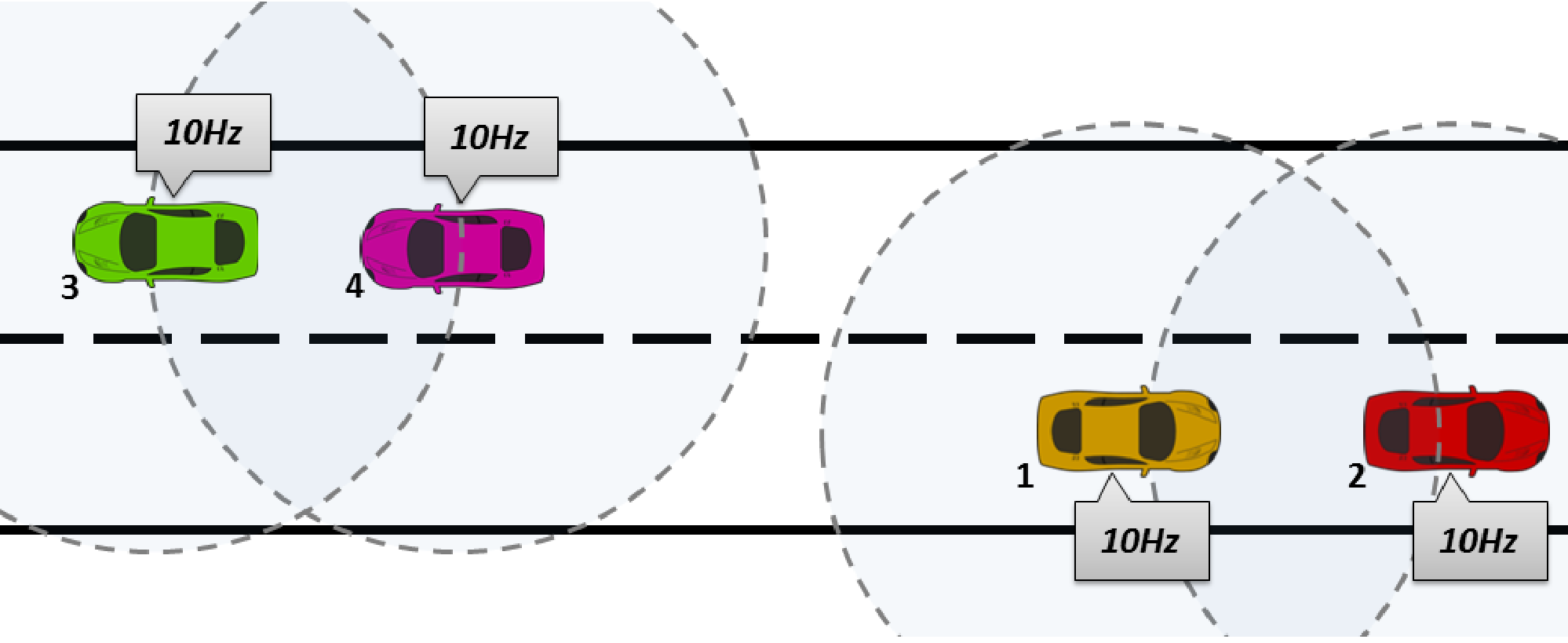}

\small{c) {No beacon rate adaptation is needed.}}
\end{center}


\vspace{.2cm}
\caption{VANET in which nodes apply FBR.}
\vspace{-.4cm}
\label{fig:ejemplo2}
\vspace{-.2cm}
\end{figure}

Figure~\ref{fig:ejemplo2} illustrates a simple example to show the behavior of VANET nodes when they adapt their beacon rates by applying FBR. 
The main features of this VANET are the same as the ones presented above.
%
There are two clusters of cars: \emph{Group 1} comprising the cars 1 and 2, that travel from left to right, and \emph{Group 2} (cars 3 and 4), that move in the opposite direction.
We define \emph{cluster of cars} as the set of VANET nodes
in which all the nodes are at least covered by the communication range of one of the other nodes of the cluster.

Initially (see Figure~\ref{fig:ejemplo2}.a), the two groups of nodes define two different clusters. All the nodes can broadcast beacons at their maximum frequency (10~$Hz$) because
the sum of the beacon rates of the nodes in the same communication range (10$+$10=20~$Hz$) does not exceed the maximum beacon rate of the channel (24~$Hz$).
After a given time (see Figure~\ref{fig:ejemplo2}.b), both groups of cars define a single cluster. 
Thus, they are aware that there are other nodes trying to broadcast beacons. 
For this reason, they have to adapt their beacon rate (FBR) to avoid network congestion because the sum of their beacon rates (10$+$10$+$10$+$10=40~$Hz$) exceeds 24~$Hz$.
Therefore, they change (adapt) their beacon frequency from 10 to 6~$Hz$ to maintain the channel load under the defined threshold (24~$Hz$).
Finally, the cars break up the cluster and build two different ones (see Figure~\ref{fig:ejemplo2}.c) like in the first case.
As a result, the network status is similar and the nodes can once again broadcast beacons at the maximum frequency.


\vspace{-0.05cm}
\section{Fair Beacon Rate Optimization Problem}
\label{sec:optimization-problem-definition}
\vspace{-0.05cm}

This section presents the formulation of the optimization problem of computing beacon rates to allow efficient CVS. 
This problem is an extension of the one presented in~\cite{Toutouh2016}.

The information required to adapt the beacon rates to the network status is the current network load (channel occupancy).
In this study, the analysis of the channel occupancy is carried out by monitoring the length of the queues in a given time window.

The FBR computation problem considers:
\begin{itemize}
\item The \emph{maximum allowed channel occupancy} \mbox{($MaxQ$ $\in$ $\mathbb{Z}$)}.
$MaxQ$ in practice represents the maximum value of queues length, i.e., the number of beacons that can be in the queue without representing a network overload (congestion).
\item A \emph{threshold limit ratio} over the maximum channel occupancy \mbox{$\alpha$ $\in$ $\left[0,1\right]$ $\subset$ $\mathbb{R}$}.
If the queue lengths exceed the \emph{effective capacity of the channel} \mbox{$\omega$ $\in$ $\left[0,MaxQ\right]$ $\subset$ $\mathbb{R}$}, which is computed according to $\omega$ = $\alpha$ $\cdot$ $MaxQ$, the protocol considers that the current network load could lead to
a congestion situation causing a degradation in performance. 
\item A \emph{set of allowed beacon rates} $BR$=$\{br^1, br^2, ..., br^k\}$. It contains all the beacon rate values ($br^i$ $\in$ $\mathbb{Z}$) that can be selected by the nodes according to the VANET application restrictions.
\item Given a vehicle $v$ that belongs to the VANET, the $NN(v)$ function returns the set that contains all the nodes inside its network coverage (1-hop neighbor nodes).

\item The \emph{occupancy of the communication channel} \mbox{$\eta(v)$ $\in$ $\left[0,100\right]$ $\subset$ $\mathbb{R}$} is computed by each node $v$ according to the ratio between its queue size and the $MaxQ$ in terms of percentage. In Eq.~(\ref{eq:channel-occupancy}), $rbr_{j}$ represents the number of received beacons from the neighbor node $j$ and $br_{v}$ is the number of beacons to be sent by $v$ during the current time window.

\begin{align}
\small
\eta(v) = \frac{\left(\sum_{j}^{NN(v)} rbr_{j} \right) + br_{v}}{MaxQ} \cdot 100\%
\label{eq:channel-occupancy}
\end{align}

\item The \emph{network balance} or \emph{fairness} \mbox{$\sigma(v)$ $\in$ $[0,+\infty)$ $\subset$ $\mathbb{R}$} is measured by using the coefficient of the variation of the beacon rates inside the neighborhood of $v$, see Eq.~(\ref{eq:fairness}). 
However, the preliminary version of the FBR optimization problem evaluates $\sigma(v)$ by means of the standard deviation~\cite{Toutouh2016}. The main reason for this change is that the standard deviation for instances with different traffic densities (different beacon rates) provides highly different results, even if the performance of the algorithms is similar. 
\begin{align}
\small
\sigma(v) = \frac{\sum\limits_{j}^{NN(v)} (br_{j} - \bar{br_{v}})^2 + (br_{v}- \bar{br_{v}})^2}{\left |NN(v)\right |}\cdot\frac{1}{\bar{br_{v}}} \label{eq:fairness}
\end{align}

$\bar{br_{v}}$ represents the average beacon data rates in the neighborhood of $v$, which is defined in Eq.~(\ref{eq:average_datarates}).
\begin{equation} \label{eq:average_datarates}
\bar{br_{v}} = \frac{\left(\sum_{j}^{NN(v)} br_{j} \right) + br_{v}}{\left |NN(v)\right |+1}
\end{equation}

\end{itemize}

The FBR optimization consist in finding the largest $br_{v} \in BR$ for each node $v$ that maximizes $\eta(v)$ (the communication channel occupancy) and minimizes $\sigma(v)$ (i.e., maximizes the network balance).
Furthermore, the computed beacon rate $br_{v}$ should not generate network congestion, i.e., $\eta(v) \leq \omega$.

In this study, the VANET nodes can generate, at least, $br^{MIN} \in BR$, which is the minimum beacon rate ($br^{MIN} < br^i,\ br^{MIN}\ \forall br^i \in BR,\ br^i \neq br^{MIN}$), but never more than $br^{MAX} \in BR$, which is the maximum  beacon rate \mbox{($br^{MAX} > br^i \ \forall br^i \in BR,\ br^i \neq br^{MIN}$)}.

\section{Swarm FREDY}
\label{sec:fredy-definition}

The Swarm FREDY \emph{(Fair beacon Rate greEDY)} congestion control method
devised in this study dynamically computes efficient beacon rates to address the FBR optimization problem proposed in the previous section.
It is fully distributed (executed individually by each VANET node), thus, no central entity is used.
In common with other swarm intelligence algorithms, the nodes perform the computations according to their \emph{own experience} and the \emph{experience of their neighbors}~\cite{Bonabeau1999}.
This section, first presents a global view of this algorithm and, then describes how the Swarm FREDY operates.

\subsection{Method Overview}
\label{sec:swarmfredy-overview}

In common with most of congestion control methods proposed in the literature,
Swarm FREDY performs two main operations: network monitoring and network components re-configuration~\cite{Lochert2007}.
In this case, the network monitoring is performed by analyzing the queues in a given time window.
Swarm FREDY applies a swarm intelligence procedure to improve the knowledge about the general network status.
%
%

The main idea consists in fairly dividing (sharing) the capacity of the channel among all the nodes in the neighborhood. Therefore, the algorithm has to evaluate the protocol queues to get the neighborhood size in order to compute an efficient \emph{desired beacon rate (DBR)}.
This can be seen as the \emph{own experience} of an individual of the swarm.
At the same time,
the VANET nodes share their computed DBR with the neighborhood to request them to change their beacon rates to the same DBR (i.e., sharing \emph{neighborhood experience}). 
The Swarm FREDY operation is described bellow. 

\begin{figure}[!h]
\setlength{\belowcaptionskip}{0.0pt}
\centering
\includegraphics[width=0.4\linewidth]{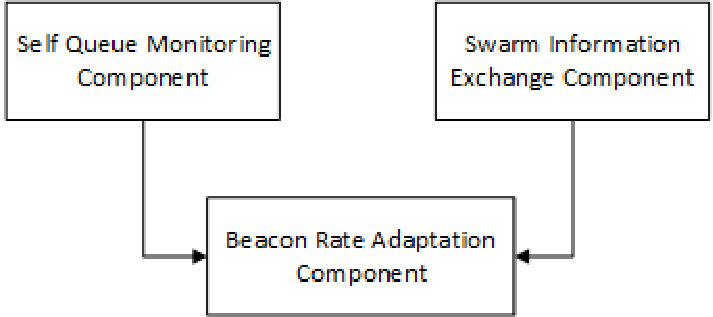}
\caption{Swarm FREDY main software components.}
\label{fig:Swarm-FREEDY-Architecture}
\vspace{-.2cm}
\end{figure}

Swarm FREDY has three main software components (see Figure~\ref{fig:Swarm-FREEDY-Architecture}):
\begin{itemize}
\item \emph{Self Queue Monitoring Component (SQMC)}, which evaluates the IEEE 802.11p protocol queues.
\item \emph{Swarm Information Exchange Component (SIEC)}, which decodes the information encoded in the received beacons.
\item \emph{Beacon Rate Adaptation Component (BRAC)}, which analyzes the information obtained by SQMC and SIEC to compute the new beacon rate.
\end{itemize}

\subsection{Swarm FREDY Operation}
\label{sec:swarmfredy-operation}
\vspace{-0.cm}

\begin{figure*}[!h]
    \centering
   \includegraphics[width=0.55\textwidth]{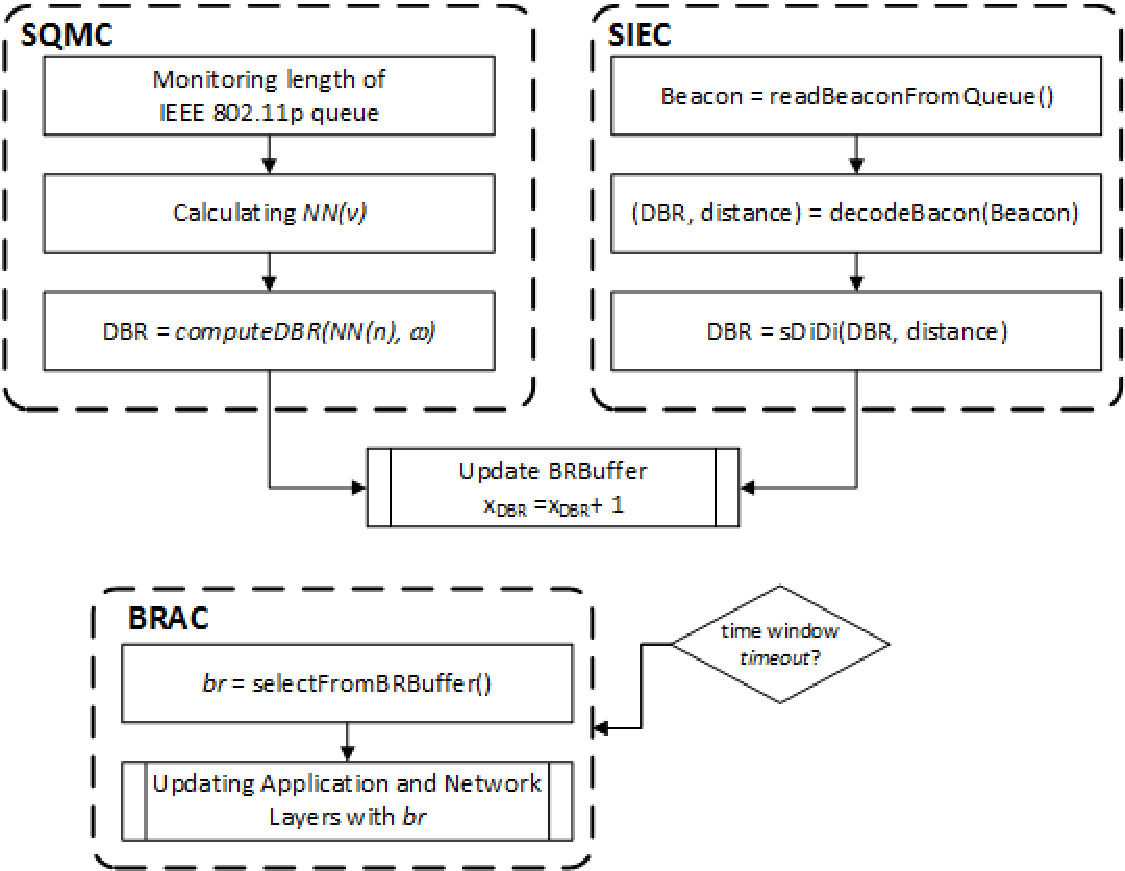}
    \caption{Complete flowchart of the Swarm FREDY algorithm.} \label{fig:swarm-procedure}
    \vspace{-0.1cm}
\end{figure*}

Swarm FREDY takes into account congestion control information received from SQMC and SIEC, which are executed permanently in parallel. This information (DBR) is stored in a temporal beacon rate buffer named (\textit{BRBuffer}) in order to utilize it in the near future computations are carried out by BRAC. Figure~\ref{fig:swarm-procedure} summarizes Swarm FREDY operation.

The BRBuffer of each node is a vector of natural values with $|BR|$=$k$ components \mbox{[$x_1$ $x_2$ ... $x_{k}$]}.
Each of the $k$ components ($x_i$) stores the number of requests received by the node to change its beacon rate to $i$ beacons per second (from itself, i.e., SQMC, or from the neighborhood, i.e., SIEC).
For example, if BRBuffer=\mbox{[0 0 10 0 25 0 0 0 0 0]}, then it means that the node has received 10 requests to change the beacon rate to 3~$Hz$ ($x_3=10$) and 25 to change to 5~$Hz$ ($x_5=25$). %

The SQMC procedure starts by analyzing the beacons received in the queues to compute the neighborhood size $|NN(v)|$.
Then, the \emph{tentative desired beacon rate (tDBR)} is computed
according to Eq.~(\ref{eq:tentative-br}).
The DBR is obtained by bounding tDBR between $br^{MIN}$ and $br^{MAX}$, as shown in Eq.~(\ref{eq:LFBR}).
The BRBuffer is updated by increasing the DBR-th component ($x_{DBR} = x_{DBR} + 1$).
Finally, the current DBR is included in the beacons to be broadcast to request the neighborhood to change their beacon rates.
\begin{equation}
tDBR = \left \lfloor \frac{\omega}{|NN(v)| + 1} \right \rfloor
\label{eq:tentative-br}
\end{equation}
\begin{equation}
DBR= \begin{cases} br^{MIN} &\mbox{if } tDBR < br^{MIN}\\
            tDBR &\mbox{if }  br^{MIN} \leq tDBR \leq br^{MAX} \\
             br^{MAX} &\mbox{if } tDBR > br^{MAX}\\
             \end{cases}
\label{eq:LFBR}
\end{equation}

Simultaneously, the SIEC procedure decodes the DBR from the beacons received and
updates the BRBuffer
according to the \emph{stochastic Distance Discriminant (sDiDi)} procedure.

\begin{figure}[!h]
\vspace{-0.1cm}
    \centering
    \includegraphics[width=0.4\textwidth]{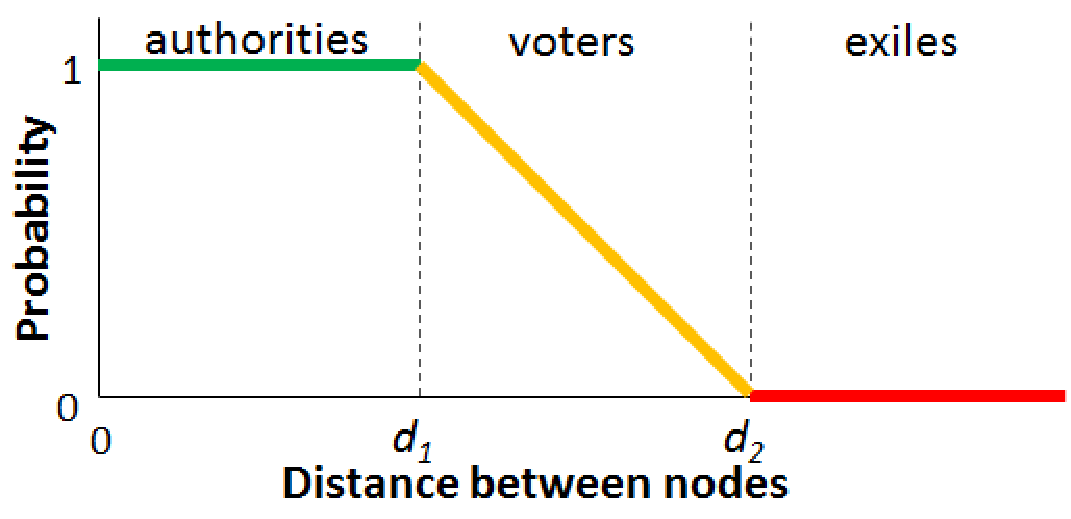}
    \caption{The sDiDi probability of including the received DBR according to the distance (node categories).} \label{fig:prob-distance}
    \vspace{-0.1cm}
    \end{figure}

The sDiDi procedure divides the neighborhood into three different categories depending on how far away they are:
i) the \emph{authorities}, which are the closest (distance $<$ $d_1$), ii) the \emph{exiles}, which are the furthest (distance $>$ $d_2$), and iii) the \emph{voters}, which are between the authorities and the exiles ($d_1$ $\leq$ distance $\leq$ $d_2$).
Therefore, if the DBR is received from an authority node, then it is always included in the BRBuffer.
If the source node of the beacon is an exiled one, then it is not included in the BRBuffer.
Finally, Eq.~(\ref{eq:voter}) defines the probability $p_i$ of including the voter's DBR in the BRBuffer, which depends on the distance between the two nodes. 
Figure~\ref{fig:prob-distance} illustrates the relationship between the nodes' distance 
and the probability of accepting the received DBR.
The DBR information is included in the buffer by increasing the DBR-th component.
\begin{equation}\label{eq:voter}
  p_i = \frac{d_2 - distance}{d2 - d1}
\end{equation}

After a given time window, BRAC is run to compute the new beaconing frequency $br^{t+1}$ according to the BRBuffer.
Thus, $br^{t+1}$ is the most requested DBR (the mode) stored in the BRBuffer.
For example, if \mbox{BRBuffer=[0 0 5 7 9 10 6 5 5 0]}, it holds that $br^{t+1}$=6 (maximum of BRBuffer is $x_6$=10).

\vspace{-0.1cm}
\section{Experimental Evaluation}
\label{sec:evaluation}
\vspace{-0.1cm}

This section presents the framework defined to evaluate the proposed Swarm FREDY by means of simulations.
The simulation environment has been defined using MATLAB.
The experimental analysis has been carried out in a Magni-Core cluster with 48 cores at 2.2 GHz, and with 48 GB RAM.
In the following subsections, we describe the congestion control methods compared,
the VANET scenarios,
and the experiments performed.

\subsection{Evaluated Methods}
\label{sec:compared-methods}

In a previous study, we evaluated nine methods to address congestion in VANET CVS applications~\cite{Toutouh2016}.
The most competitive one was Swarm DIFRA, which estimates channel load in a distributed manner and dynamically adapts the beacon rate of each node by using deterministic computations. %
Thus, as it was the most competitive congestion control method, 
we decided to 
compare it with our present approach.

The performance of Swarm FREDY is highly dependent on the $d_1$ and $d_2$ values used in the sDiDi method to discriminate the information obtained from the received beacons (see Section~\ref{sec:swarmfredy-operation}).
We therefore evaluate Swarm FREDY using different ($d_1$,$d_2$) configurations.
The selected distances are multiples of 50~meters and they range from 0 to 250~meters. Table~\ref{tab:d1d2-configurations} shows the 15~Swarm FREDY configurations analyzed in this study.
The time window for the execution of BRAC procedure is set to 1~second.

\begin{table}[!h]
\centering
\caption{Swarm FREDY configurations analyzed.}\label{tab:d1d2-configurations}
\setlength{\tabcolsep}{3pt}
\small
\begin{tabular}{crrrrrr}
  &&\multicolumn{5}{c}{$d_2$} \\
  & & \textbf{50} & \textbf{100} & \textbf{150} & \textbf{200} & \textbf{250} \\
  \cline{3-7}
  \multirow{5}{*}{$d_1$} & \multicolumn{1}{r|}{\textbf{0}} & (0,50) & (0,100) & (0,150) & (0,200) & (0,250) \\
  &\multicolumn{1}{r|}{\textbf{50}} & - & (50,100)  & (50,150)  & (50,200)   & (50,250)  \\
  &\multicolumn{1}{r|}{\textbf{100}}   & - & - & (100,150) & (100,200) &  (150,250)\\
  &\multicolumn{1}{r|}{\textbf{150}}  & - & - & - &  (150,200) &  (150,250) \\
  &\multicolumn{1}{r|}{\textbf{200}}   & -& -& - & - &  (200,250) \\
  \cline{3-7}
\end{tabular}
\end{table}

\subsection{VANET Scenarios}
\label{sec:road-traffic}

The methods under comparison are studied in a road that covers ten kilometers long and has six lanes wide (three lanes in each direction) for 150~seconds.
In order to evaluate their robustness over different road traffic situations, seven VANET scenarios are defined by changing the road traffic density (number of moving vehicles): 500, 750, 1000, 1250, 1500, 1750, and 2000.

A realistic mobility model based on Intelligent-Driver Model (IDM) has been defined~\citep{Lan2008}. 
The vehicles are initially assigned to a given lane randomly, 
with a higher probability of being assigned to the outer lanes than the inner ones, as in real world roads.
%
The speed of the vehicles is higher in the external lanes than in the internal ones. 
The distances between vehicles and the speeds are computed according to \emph{the square law}, i.e., $speed^2 \simeq distance/100$~\cite{Toutouh2016}.
As the initial location of the vehicles on the roads is non-deterministic, each time a simulation is run a different traffic scenario is generated (even if they have the same number of vehicles). 
This gives our study a better background by considering a huge number of realistic VANET scenarios that represent different real world road traffic situations.

In order to model the communications by using IEEE 802.11p, the probabilistic Three-Log Distance propagation model is used with 5.8~GHz radio operating at 6~Mbps data rate.
The communication range of the radio devices $r$ is set to 250~meters.
The maximum size of the IEEE 802.11p queues ($MaxQ$) is 400 beacons and threshold limit ratio ($\alpha$) is 0.8.
Finally, the CVS applications running in each vehicle require beacons of 100~bytes to be exchanged with a frequencies that range from 1~$Hz$ to 10~$Hz$.

\subsection{Design of Experiments}
\label{sec:defined-experiments}

In this study, although we focus our analysis on the two metrics evaluated by the FBR optimization problem: the \emph{channel occupancy (usage)} and the \emph{network balance}, another two metrics to evaluate the QoS of the CVS applications are also considered: the \emph{individual beacon rate}, which is the average beacon rate of the nodes during the simulation, and the \emph{beacon rate stability}, which represents the number of times that a node has to re-adapt its beacon rate to the current network status.

The individual beacon rate evaluates the reliability of CVS applications, as introduced in Section~\ref{sec:BF-QoS}. The stability evaluates the number of beacon rate changes performed during the simulation  because it represents the stability of the QoS provided by CVS applications.

As the road traffic and the communications are non-deterministic,
each analyzed congestion control method is simulated 50~times over the same VANET scenario
because each simulation produces different results. 
Thus, to determine the significance of the comparison, statistical tests are applied (Kolmogorov-Smirnov and Aligned Friedman Rank tests)~\cite{sheskin2003} with a confidence level of 99\% (\emph{p-value}$<$0.01).

%

\section{Numerical Results}
\label{sec:experimental-results}

This section summarizes and analyzes the main results of the experimental evaluation of the 15~variants of our Swarm FREDY analyzed in this study. 
Additionally, the experiments includes Swarm DIFRA congestion control methods as a baseline for comparison purposes because 
it has already demonstrate more competitive performance than other well-known methods in previous studies~\cite{Toutouh2016}. 
In this way, we simplify the experimental analysis by avoiding including all these other well-known classic congestion control methods.  
The result distributions computed are not normal according to the Kolmogorov-Smirnov statistical test.
Therefore, we show the median over the 50~independent simulations 
for each scenario. 
In order to simplify the tables of results the Swarm FREDY methods are named SF($d_1$,$d_2$) and Swarm DIFRA is identified by SD. 
%

For each evaluated metric, a boxplot graph of the results of the most representative VANET scenario (road traffic) of the global behavior is shown, aiming at improving the comprehension of the obtained results.

\subsection{Individual Beacon Rate}
\label{sec:br-result}

\begin{figure}[!ht]
	\vspace{-0.7cm}
    \centering
    \includegraphics[width=0.45\textwidth]{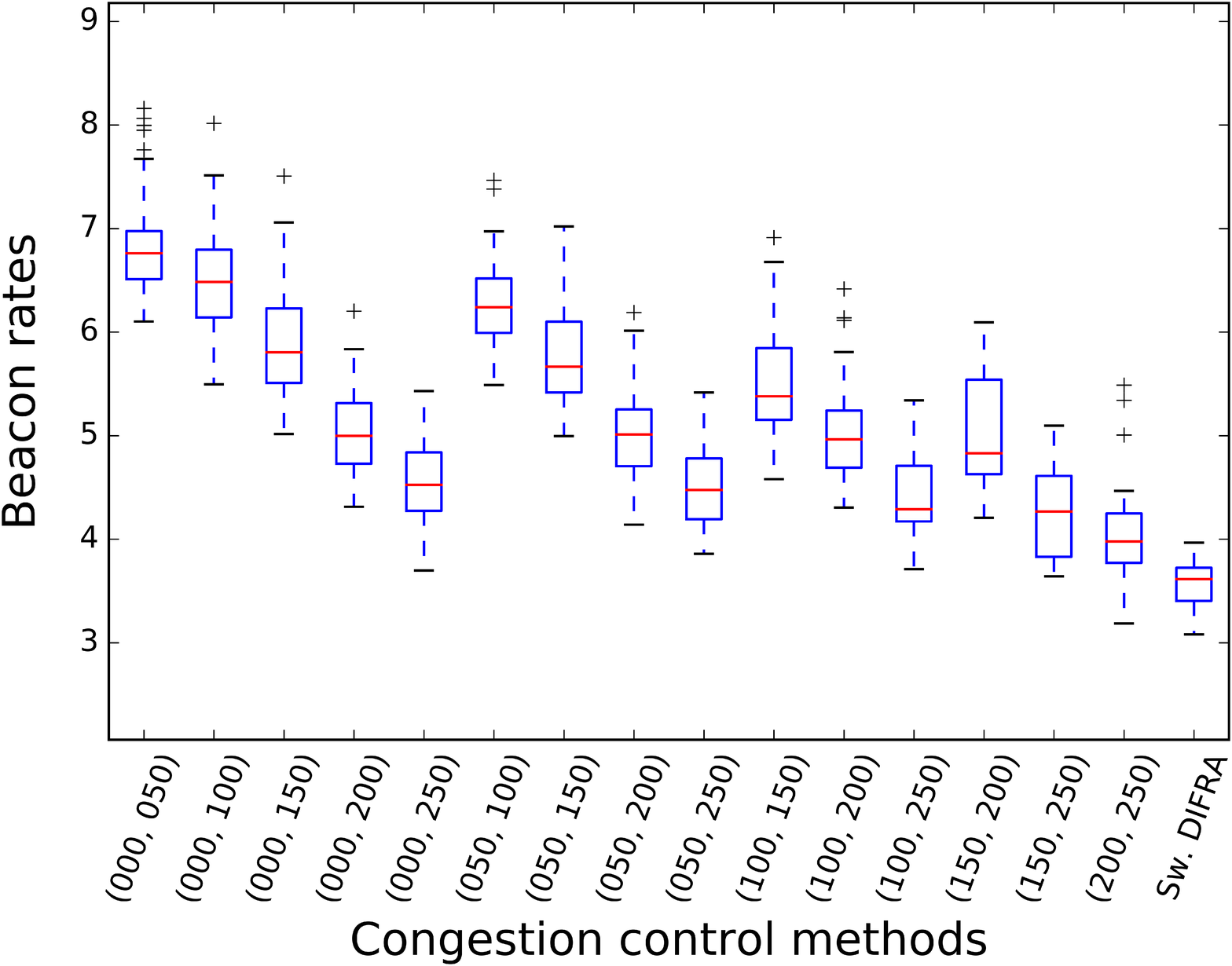}
    \caption{Vehicles beacon rates results for the scenario with 1000 vehicles.} \label{fig:br}
\end{figure}

Table~\ref{tab:br-median} summarizes the individual beacon rates by showing the median values over the 50~simulations.
Figure~\ref{fig:br} shows the beacon rates used by the mobile nodes for the scenarios with 1000~vehicles.
Finally, Table~\ref{tab:br-friedman} summarizes the Aligned Friedman Rank results for this metric (\emph{p-value}$<$0.01).

\begin{table}[!h]
\centering
\caption{Median results of the vehicles' beacon rates (Hz or beacons per second) for each VANET scenario and congestion control method.}
\setlength{\tabcolsep}{3.5pt}
\renewcommand{\arraystretch}{0.95}
\small
\begin{tabular}{lrrrrrrr}
\hline		
\textbf{Method} & \textbf{500$\,$veh.} & \textbf{750$\,$veh.} & \textbf{1000$\,$veh.} & \textbf{1250$\,$veh.} & \textbf{1500$\,$veh.} & \textbf{1750$\,$veh.} & \textbf{2000$\,$veh.}\\
\hline		
SF(000,050) & \textbf{7.251} 	 & \textbf{6.125} 	 & \textbf{5.174} 	 & \textbf{4.678} 	 & \textbf{3.939} 	 & 3.339 	 & 2.999 	\\
SF(000,100) & 7.208  	 & 6.044 	 & 5.100 	 & 4.662 	 & 3.930 	 & \textbf{3.348} 	 & 2.996 	\\
SF(000,150) & 7.208 	 & 5.932 	 & 5.046 	 & 4.645 	 & 3.905 	 & 3.337 	 & 2.997 	\\
SF(000,200) & 7.139 	 & 5.898 	 & 4.987 	 & 4.611 	 & 3.902 	 & 3.309 	 & 2.982 	\\
SF(000,250) & 7.083 	 & 5.825 	 & 4.944 	 & 4.601 	 & 3.862 	 & 3.304 	 & 2.978 	\\
SF(050,100) & 7.234 	 & 6.018 	 & 5.113 	 & 4.632 	 & 3.926 	 & 3.324 	 & \textbf{3.003} 	\\
SF(050,150) & 7.158 	 & 5.951 	 & 5.041 	 & 4.630 	 & 3.893 	 & 3.312 	 & 2.975 	\\
SF(050,200) & 7.215 	 & 5.886 	 & 5.013 	 & 4.603 	 & 3.890 	 & 3.316 	 & 2.996 	\\
SF(050,250) & 7.126 	 & 5.804 	 & 4.945 	 & 4.561 	 & 3.866 	 & 3.297 	 & 2.962 	\\
SF(100,150) & 7.231 	 & 5.908 	 & 5.011 	 & 4.594 	 & 3.894 	 & 3.313 	 & 2.988 	\\
SF(100,200) & 7.090 	 & 5.871 	 & 4.944 	 & 4.584 	 & 3.861 	 & 3.307 	 & 2.956 	\\
SF(100,250) & 7.082 	 & 5.771 	 & 4.941 	 & 4.560 	 & 3.850 	 & 3.294 	 & 2.963 	\\
SF(150,200) & 7.112 	 & 5.874 	 & 4.962 	 & 4.583 	 & 3.856 	 & 3.286 	 & 2.978 	\\
SF(150,250) & 7.058 	 & 5.747 	 & 4.940 	 & 4.547 	 & 3.849 	 & 3.271 	 & 2.956 	\\
SF(200,250) & 7.036 	 & 5.778 	 & 4.857 	 & 4.549 	 & 3.826 	 & 3.267 	 & 2.949 	\\
SD & 7.025 	 & 5.748 	 & 4.879 	 & 4.512 	 & 3.808 	 & 3.242 	 & 2.935 	\\
\hline		
\end{tabular}
\label{tab:br-median}
\end{table}

\begin{table}[!h]
\centering
\vspace{-0.2cm}
\renewcommand{\arraystretch}{0.95}
\caption{Aligned Friedman Ranking of the vehicles' beacon rates results.}
\small
\begin{tabular}{lrr}
\hline		
\textbf{Congestion control method}  & \textbf{Ranking position} & \textbf{Rank value}\\
\hline		
SF(050,100) & 1 & 3075.158 \\
SF(000,050) & 2 & 2954.868 \\
SF(000,100) & 3 & 2870.132 \\
SF(100,150) & 4 & 2620.711 \\
SF(000,150) & 5 & 2549.053 \\
SF(050,200) & 6 & 2508.342 \\
SF(050,150) & 7 & 2296.500 \\
SF(000,200) & 8 & 2104.737 \\
SF(050,250) & 9 & 1955.895 \\
SF(150,200) & 10 & 1868.947 \\
SF(100,200) & 11 & 1770.026 \\
SF(000,250) & 12 & 1663.737 \\
SF(100,250) & 13 & 1537.737 \\
SF(200,250) & 14 & 1522.263 \\
SF(150,250) & 15 & 1491.868 \\
SD & 16 & 1266.026 \\
\hline		
\multicolumn{3}{c}{\emph{p-value} $<<$ 0.0000001} \\
\hline		
\end{tabular}
\label{tab:br-friedman}
\vspace{-0.3cm}
\end{table}

As expected, the higher the road traffic density (number of vehicles on the road) the lower the beacon rates. 
This common sense result has been achieved automatically (an important feature of our simulation platform). 
In these experiments the beacon rates decrease from higher than 7~Hz to lower than 3~Hz.
This illustrates the variations in the communications depending on the road traffic density and the importance of efficiently managing the channel usage.

According to the results in Tables~\ref{tab:br-median} and~\ref{tab:br-friedman}, Swarm DIFRA is statistically the least competitive method evaluated, since it provides the lowest beacon rates  (see Figure~\ref{fig:br}). 
In those scenarios with lower road traffic and where the nodes are able to exchange information with higher beacon rates, the differences between our approach and Swarm DIFRA is higher.

Let us analyze the different variants of Swarm FREDY. 
We observe in Figure~\ref{fig:br} that the shorter the $d_2$ distance the higher the beacon rate.
This is because the communication channel is divided into a lower number of considered nodes (i.e., authorities and voters), since the algorithm classifies nodes from shorter distances as exiles.

The three most competitive Swarm FREDY configurations are those that do not use data packets from further than 100~m for the beacon rate computations. Specifically, the best performance of this metric is achieved by the protocol configuration $d_1$=50~m and $d_2$=100~m. 

The least competitive Swarm FREDY configurations for this metric (lower beacon rates) are those that take into account beacons from the farthest nodes ($d_2$=250~m). Indeed, their performance for this metric is close to that of Swarm DIFRA.

Summarizing, Swarm FREDY methods allow communications with higher beacon rates than Swarm DIFRA, i.e., the reliability of CVS applications is increased, and therefore, the road traffic safety is improved.
This improvement provided by Swarm FREDY is higher for road traffic scenarios with lower traffic densities, where the vehicles are moving at faster speeds, and therefore, the CVS applications require a higher information refresh rate by exchanging beacons with higher data rates to avoid hazardous situations.

%

\subsection{Channel Usage}
\label{sec:cu-results}

We now present the channel usage results in terms of the percentage of protocol queue filling. 
This metric is evaluated because it is not only affected by the communication beacon rate in the neighborhood, but also by other factors such as signal propagation and packet collisions.
In terms of channel usage, the higher the better. 

According to Table~\ref{tab:cu-median}, 
there is no clear trend for the channel occupancy and the road traffic density. The scenarios with the three largest channel occupancies are those with 750, 1750, and 2000~vehicles.

In contrast to the individual beacon rate, in which Swarm FREDY (0,50) presented the most competitive results for most scenarios, for this metric (channel usage), 
there is not a specific Swarm FREDY configuration that clearly stands out from the others. 
Thus, Swarm FREDY is robust for this metric.

\begin{figure}[!h]
\vspace{-0.2cm}
    \centering
    \includegraphics[width=0.45\textwidth]{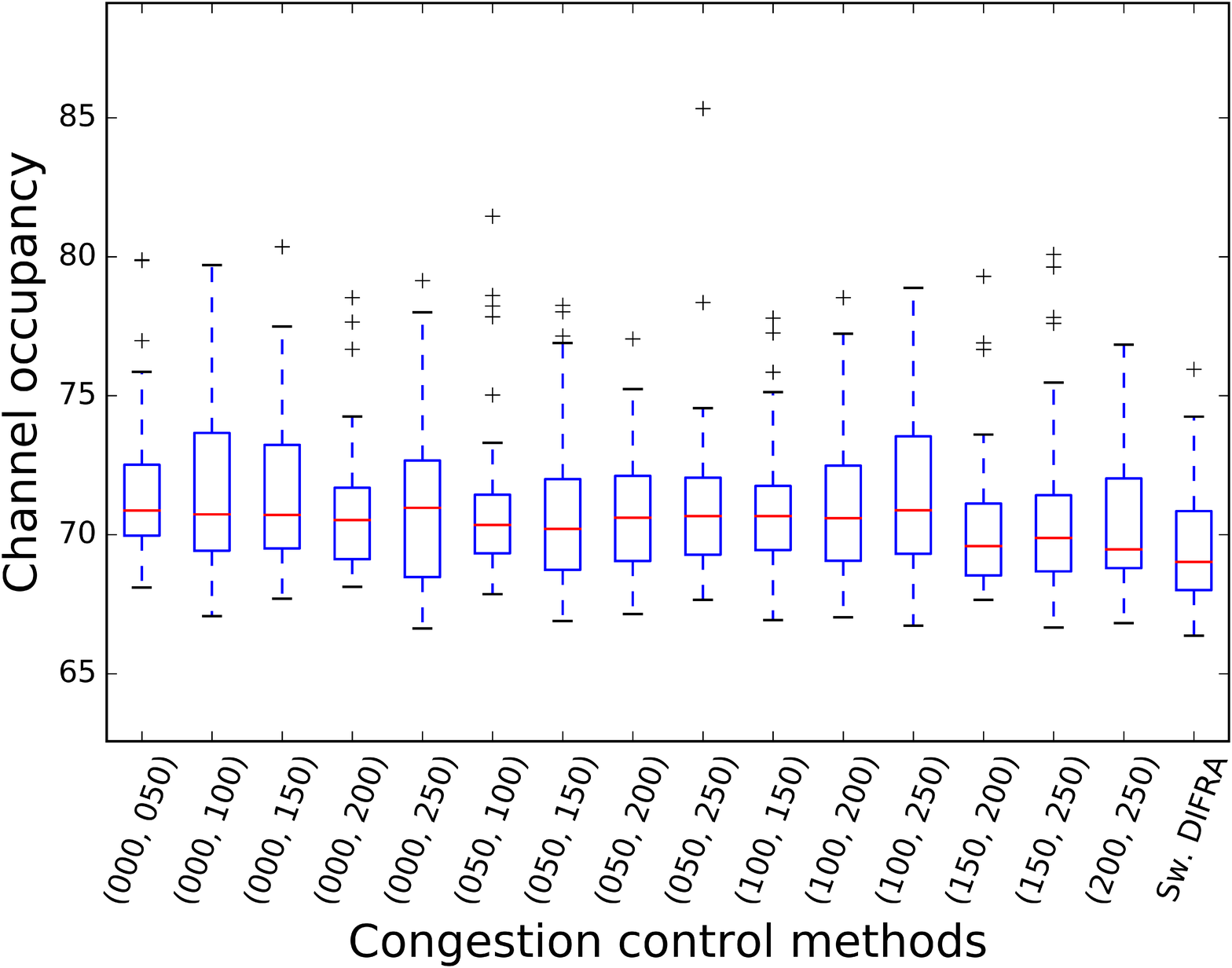}
    \caption{Channel occupancy results (in \%) for the scenario with 1750 vehicles.} \label{fig:cu}
\end{figure}

The lowest channel occupancy is obtained by Swarm DIFRA.
This can be seen in Figure~\ref{fig:cu} for the scenario with 1750~vehicles.
The main reason for this result is that when the VANET nodes use this algorithm, they communicate with the lowest beacon rates (as shown in Section~\ref{sec:br-result}), and therefore, they use the available shared medium less efficiently.

\begin{table}[!h]
\caption{Median results of the channel usage for each VANET scenario and congestion control method.} \label{tab:cu-median}
\centering
\setlength{\tabcolsep}{3.5pt}
\renewcommand{\arraystretch}{0.94}
\small
\begin{tabular}{lrrrrrrr}
\hline		
\textbf{Method} & \textbf{500$\,$veh.} & \textbf{750$\,$veh.} & \textbf{1000$\,$veh.} & \textbf{1250$\,$veh.} & \textbf{1500$\,$veh.} & \textbf{1750$\,$veh.} & \textbf{2000$\,$veh.}\\
\hline		
SF(000,050) & 66.770 & 73.124 & \textbf{66.844} & 65.780 & 69.835 & 70.917 & 76.181 \\
SF(000,100) & 66.888 & \textbf{73.430} & 65.774 & \textbf{65.956} & 69.967 & 70.808 & 76.676 \\
SF(000,150) & 66.797 & 72.171 & 65.704 & 65.869 & 69.358 & 70.918 & 77.174 \\
SF(000,200) & 67.026 & 72.552 & 66.011 & 65.294 & 70.063 & 70.534 & 75.612 \\
SF(000,250) & 66.378 & 72.213 & 64.909 & 65.507 & 68.910 & \textbf{71.005} & 76.829 \\
SF(050,100) & \textbf{67.161} & 72.645 & 66.685 & 65.807 & \textbf{70.734} & 70.363 & 77.365 \\
SF(050,150) & 66.831 & 72.205 & 66.041 & 65.307 & 69.174 & 70.461 & 75.588 \\
SF(050,200) & 66.745 & 73.313 & 66.482 & 65.184 & 70.036 & 70.749 & \textbf{77.568} \\
SF(050,250) & 66.197 & 71.882 & 65.546 & 65.138 & 69.332 & 70.672 & 75.915 \\
SF(100,150) & 67.055 & 71.938 & 65.350 & 64.995 & 69.650 & 70.611 & 76.585 \\
SF(100,200) & 66.542 & 72.367 & 65.141 & 64.727 & 69.119 & 70.929 & 75.157 \\
SF(100,250) & 66.499 & 71.351 & 65.351 & 64.709 & 69.184 & 70.922 & 75.595 \\
SF(150,200) & 66.287 & 72.100 & 64.704 & 65.504 & 69.181 & 69.599 & 76.348 \\
SF(150,250) & 66.141 & 71.186 & 64.676 & 65.017 & 69.073 & 69.800 & 75.733 \\
SF(200,250) & 66.674 & 71.418 & 64.518 & 65.098 & 69.139 & 69.511 & 76.581 \\	
SD & 66.091 & 71.237 & 64.760 & 64.415 & 68.704 & 69.097 & 74.677 \\
\hline		
\end{tabular}
\end{table}

\begin{table}[!h]
\caption{Aligned Friedman Ranking of the channel usage results.}
\scriptsize
\centering
\label{tab:cu-friedman}
\small
\renewcommand{\arraystretch}{0.94}
\begin{tabular}{lrr}
\hline		
\textbf{Congestion control method}  & \textbf{Ranking position} & \textbf{Rank value}\\
\hline		
SF(050,100) & 1 & 2861.566 \\
SF(000,100) & 2 & 2483.934 \\
SF(000,150) & 3 & 2426.737 \\
SF(050,200) & 4 &2413.750 \\
SF(100,150) & 5 &2396.987 \\
SF(000,050) & 6 & 2379.395 \\
SF(000,200) & 7 & 2327.816 \\
SF(100,200) & 8 & 2300.645 \\
SF(050,150) & 9 & 2286.368 \\
SF(150,200) & 10 & 2021.750 \\
SF(100,250) & 11 & 1909.013 \\
SF(200,250) & 12 & 1839.474 \\
SF(000,250) & 13 & 1760.724 \\
SF(050,250) & 14 & 1643.474 \\
SF(150,250) & 15 & 1635.000 \\
SD & 16 & 1369.368 \\
\hline
\multicolumn{3}{c}{\emph{p-value} $<<$ 0.0000001} \\
\hline		
\end{tabular}
\vspace{-0.2cm}
\end{table}

The performance differences between Swarm DIFRA and Swarm FREDY are higher in the scenarios with higher road traffic densities (i.e., higher congestion control problems). Here, our proposed method more efficiently manages the communication channel in more complicated situations.

According to the Aligned Friedman ranking results (see Table~\ref{tab:cu-friedman}), the Swarm FREDY configurations, which use the channel less efficiently, are those with larger $d_2$ distances ($d_2$=250).
Therefore as occurs with the individual beacon rate metric, if the algorithm considers exiles from shorter distances it performs better.
Swarm FREDY (50,100) achieves significantly more competitive results.

Summarizing, Swarm FREDY maximizes the channel usage, which is one of the objectives of the FBR optimization problem. The CVS applications exchange larger amounts of data, while avoiding channel congestion.
Thus, the likelihood of exchanging the information required by the CVS applications increases, providing safer road journeys. These results answer \textbf{RQ1} because it can be seen that an efficient congestion control method is important to improve road safety and traffic efficiency.

%
%
%
%
%

\vspace{-0.15cm}
\subsection{Network Balance}
\label{sec:nb-results}
\vspace{-0.05cm}

The network balance is evaluated in terms of $\sigma(v)$.
The congestion control methods that allow higher fairness 
provide 
closer beacon rates among the VANET nodes. 
According to Eq.~(\ref{eq:fairness}), the lower $\sigma(v)$ the better the balance.
Table~\ref{tab:nb-median} shows the median results for the 50~simulations for each scenario.
There is no clear trend in these results, but in general when the road traffic density grows, the network balance worsens ($\sigma(v)$ increases).
This is mainly because the coefficient of variation increases when the values of the distribution (i.e., beacon rates of the nodes) is lower.

\begin{table}[!h]
\caption{Median results of the network balance for each VANET scenario and congestion control method.} \label{tab:nb-median}
\centering
\renewcommand{\arraystretch}{0.95}
\setlength{\tabcolsep}{3.5pt}
\small\begin{tabular}{lrrrrrrr}
\hline		
\textbf{Method} & \textbf{500$\,$veh.} & \textbf{750$\,$veh.} & \textbf{1000$\,$veh.} & \textbf{1250$\,$veh.} & \textbf{1500$\,$veh.} & \textbf{1750$\,$veh.} & \textbf{2000$\,$veh.}\\
\hline		
SF(000,050) & 0.476  & 0.525  & 0.677  & 0.769  & 0.643  & 0.521  & 0.434   \\
SF(000,100) & 0.457  & 0.504  & 0.647  & 0.708  & 0.599  & 0.509 	& 0.400   \\
SF(000,150) & 0.417  & 0.461  & 0.577  & 0.676  & 0.563  & 0.455  & 0.393   \\
SF(000,200) & 0.366  & 0.409  & 0.497  & 0.633  & 0.521  & 0.441  & 0.396   \\
SF(000,250) & 0.316  & 0.369  & 0.452  & 0.588  & 0.465  & 0.390  & 0.315   \\
SF(050,100) & 0.434  & 0.491  & 0.624  & 0.703  & 0.587  & 0.492  & 0.388   \\
SF(050,150) & 0.396  & 0.446  & 0.567  & 0.664  & 0.546  & 0.459  & 0.384   \\
SF(050,200) & 0.365  & 0.406  & 0.500  & 0.619  & 0.523  & 0.425  & 0.357   \\
SF(050,250) & 0.315  & 0.374  & 0.448  & 0.555  & 0.461  & 0.378  & 0.318   \\
SF(100,150) & 0.395  & 0.445  & 0.544  & 0.641  & 0.529  & 0.428  & 0.377   \\
SF(100,200) & 0.345  & 0.405  & 0.497  & 0.627  & 0.528  & 0.426  & 0.372   \\
SF(100,250) & 0.308  & 0.361  & 0.429  & 0.557  & 0.463  & 0.359  & 0.320   \\
SF(150,200) & 0.353  & 0.402  & 0.482  & 0.604  & 0.486  & 0.412  & 0.359   \\
SF(150,250) & 0.299  & 0.346  & 0.425  & 0.535  & 0.444  & 0.360  & 0.297   \\
SF(200,250) & 0.283  & 0.337  & 0.398  & 0.506  & 0.412  & 0.332  & 0.273   \\
SD & \textbf{0.255}  & \textbf{0.307}  & \textbf{0.361}  & \textbf{0.458}  & \textbf{0.359}  & \textbf{0.309}  & \textbf{0.255}   \\
\hline		
\end{tabular}
\vspace{-0.1cm}
\end{table}

\begin{figure}[!ht]
    \vspace{-0.5cm}
    \centering
   \includegraphics[width=0.45\textwidth]{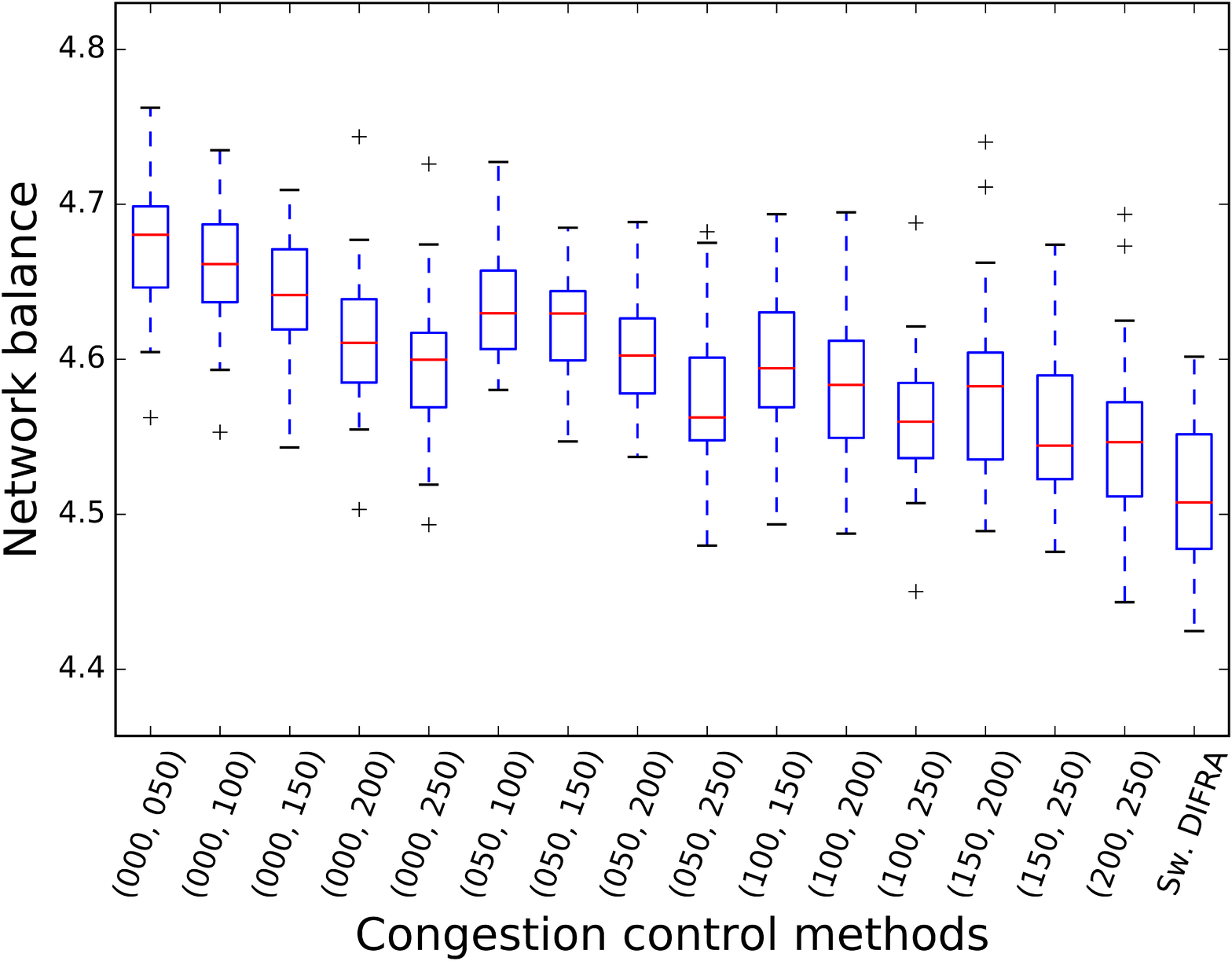}
    \caption{Network balance results for the scenario with 1250 vehicles.} \label{fig:nb}
    \vspace{-0.1cm}
\end{figure}

\begin{table}[!h]
\caption{Aligned Friedman Ranking of the network balance results.} \label{tab:nb-friedman}
\small
\renewcommand{\arraystretch}{0.95}
\centering
\begin{tabular}{lrr}
\hline		
\textbf{Congestion control method}  & \textbf{Ranking position} & \textbf{Rank value}\\
\hline		
SD & 1 & 251.773\\
SF(200,250) & 2 & 584.648\\
SF(150,250) & 3 & 912.441\\
SF(100,250) & 4 & 1150.832\\
SF(050,250) & 5 & 1312.114\\
SF(000,250) & 6 & 1417.403\\
SF(150,200) & 7 & 2076.919\\
SF(050,200) & 8 & 2272.590\\
SF(100,200) & 9 & 2299.212\\
SF(000,200) & 10 & 2489.890\\
SF(100,150) & 11 & 2782.993\\
SF(050,150) & 12 & 3020.392\\
SF(000,150) & 13 & 3212.648\\
SF(050,100) & 14 & 3540.520\\
SF(000,100) & 15 & 3662.623\\
SF(000,050) & 16 & 3965.004\\
\hline
\multicolumn{3}{c}{\emph{p-value} $<<$ 0.0000001} \\
\hline		
\end{tabular}
\vspace{-0.2cm}

\end{table}

In this case, Swarm DIFRA obtains the most competitive results (the lowest $\sigma(v)$). 
These results are statistically confirmed by the Aligned Friedman Ranking test (see Table~\ref{tab:nb-friedman}).
The main reason for this is that Swarm DIFRA does not discriminate between the neighbor nodes~\cite{Toutouh2016},
and therefore, it considers the whole neighborhood when computing the desired beacon rates.
This computes lower beacon rates for all the nodes than Swarm FREDY.
At first sight, it is fairer than Swarm FREDY but it causes a high degradation of the QoS (amount of data exchanged), and it is really bad. 

Analyzing Swarm FREDY, the results in Table~\ref{tab:nb-median} do not show a clear behavior in terms of balance for all the methods.
However, it can be seen that the longer $d_2$ the better results.
Thus, if the set of authorities and voters nodes grows Swarm FREDY obtains more balanced beacon rates.
Figure~\ref{fig:nb} illustrates this behavior for the scenario with 1250 vehicles.


\subsection{Beacon Rate Stability}
\label{sec:bradapt-results}


This section evaluates the beacon rate stability in terms of the number of beacon rate changes during the simulation,
i.e., the number of times the nodes had to modify their beacon rate. 
The lower the number of times the beacon rate changed, the more stable the broadcasting method. 
Designers are able to create more reliable VANET applications when the performance of the network is more stable and predictable, 
so we look for algorithms showing this feature.

Table~\ref{tab:bradap-median} summarizes these results 
for the 50~simulations.
When the road traffic density grows the number of beacon adaptations increases.
This is mainly because having more vehicles makes the distributed computation of efficient and accurate beacon rates more difficult.

\begin{figure}[!ht]
\vspace{-0.3cm}
    \centering
    \includegraphics[width=0.45\textwidth]{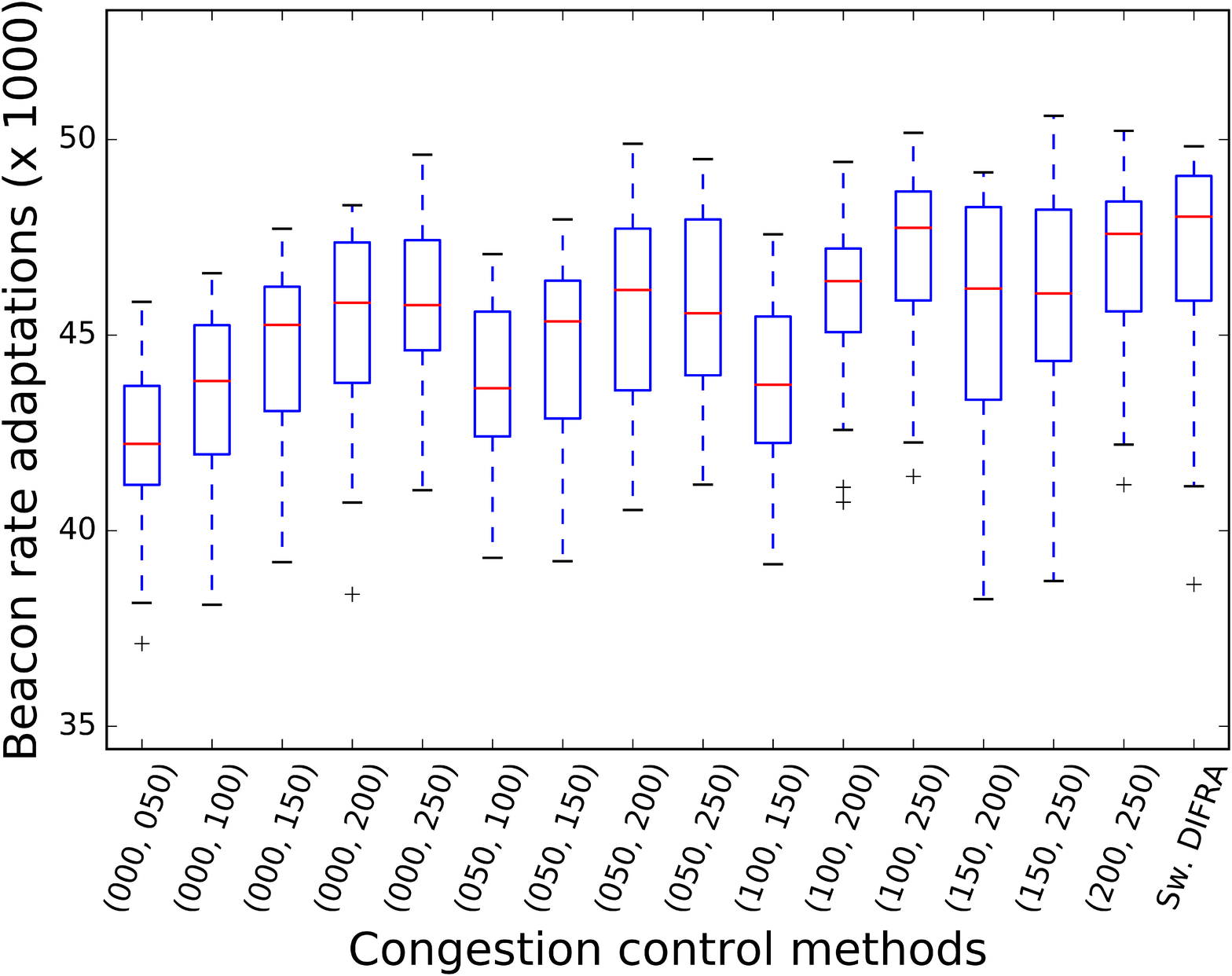}
    \vspace{-0.1cm}
    \caption{Network balance results for the scenario with 500 vehicles.}
    \label{fig:bradap}
\end{figure}

In general, for the scenarios with lower road traffic densities (number of vehicles lower or equal to 1000)
the Swarm FREDY methods provide more competitive results than Swarm DIFRA.
In these scenarios they require lower beacon rate transformations, and therefore, the QoS of the communications is preserved for longer. Figure~\ref{fig:bradap} illustrates these results.
For the scenarios with higher traffic densities our approaches are still more stable than Swarm DIFRA. 

\begin{table}[!h]
\vspace{-0.0cm}
\caption{Median results of the number of beacon rate adaptations ($\times 10^5$) for each VANET scenario.} \label{tab:bradap-median}
\centering
\renewcommand{\arraystretch}{0.95}
\setlength{\tabcolsep}{2.9pt}
\small
\begin{tabular}{lrrrrrrr}
\hline		
\textbf{Method} & \textbf{500$\,$veh.} & \textbf{750$\,$veh.} & \textbf{1000$\,$veh.} & \textbf{1250$\,$veh.} & \textbf{1500$\,$veh.} & \textbf{1750$\,$veh.} & \textbf{2000$\,$veh.}\\
\hline		
SF(000,050) & 	 \textbf{42.216} 	&	 \textbf{67.241} 	&	 \textbf{70.574} 	&	99.531	&	168.595	&	197.015	&	224.945	\\
SF(000,100) & 	43.722	&	68.996	&	71.463	&	99.036	&	168.690	&	190.676	&	225.170	\\
SF(000,150)  &	45.288	&	70.738	&	72.474	&	99.297	&	168.720	&	 \textbf{196.735} 	&	224.930	\\
SF(000,200)  &	45.837	&	71.775	&	73.354	&	100.240	&	168.765	&	196.960	&	 \textbf{224.795} 	\\
SF(000,250)  &	45.707	&	73.579	&	73.494	&	99.302	&	 \textbf{168.500} 	&	196.875	&	224.940	\\
SF(050,100)  &	43.887	&	69.070	&	71.728	&	99.496	&	168.690	&	196.875	&	225.140	\\
SF(050,150)  &	45.445	&	70.821	&	72.200	&	98.898	&	168.895	&	196.885	&	225.110	\\
SF(050,200)  &	46.204	&	71.807	&	73.081	&	98.704	&	168.535	&	196.830	&	224.970	\\
SF(050,250)  &	45.505	&	73.088	&	73.619	&	98.462	&	168.730	&	197.095	&	224.995	\\
SF(100,200) & 	46.387	&	71.780	&	73.460	&	99.030	&	168.750	&	196.845	&	225.100	\\
SF(100,250)  &	47.643	&	73.728	&	73.872	&	 \textbf{98.085} 	&	168.765	&	196.945	&	225.095	\\
SF(100,150)   &	43.706	&	71.107	&	72.525	&	100.605	&	168.670	&	196.960	&	225.150	\\
SF(150,200)  &	45.925	&	71.857	&	73.958	&	99.144	&	168.640	&	196.745	&	225.050	\\
SF(150,250) & 	46.064	&	73.484	&	74.055	&	99.662	&	168.660	&	196.965	&	224.790	\\
SF(200,250)  &	47.736	&	73.434	&	74.643	&	98.782	&	168.750	&	196.940	&	225.185	\\
SD &	48.157	&	73.701	&	74.650	&	99.965	&	168.880	&	196.900	&	224.730	\\

\hline		
\end{tabular}
\end{table}

\begin{table}[!h]
\caption{Aligned Friedman Ranking of the stability results.} \label{tab:bradap-friedman}
\centering
\renewcommand{\arraystretch}{0.95}
\small
\begin{tabular}{lrr}
\hline		
\textbf{Congestion control method}  & \textbf{Ranking position} & \textbf{Rank value}\\
\hline		
SF(000,050) & 1 & 1407.701 \\
SF(000,100) & 2 & 1589.606 \\
SF(050,100) & 3 & 1654.310 \\
SF(000,150) & 4 & 1885.114 \\
SF(050,150) & 5 & 1920.943 \\
SF(100,150) & 6 & 1935.092 \\
SF(050,200) & 7 & 2138.353 \\
SF(000,200) & 8 & 2253.535 \\
SF(150,200) & 9 & 2273.104 \\
SF(100,200) & 10 & 2338.093 \\
SF(050,250) & 11 & 2415.907 \\
SF(000,250) & 12 & 2475.255 \\
SF(150,250) & 13 & 2579.247 \\
SF(100,250) & 14 & 2615.007 \\
SD & 15 & 2719.504 \\
SF(200,250) & 16 & 2751.229 \\
\hline		
\multicolumn{3}{c}{\emph{p-value} $<<$ 0.0000001} \\
\hline		
\end{tabular}
\end{table}

The Aligned Friedman ranking statistical results are shown in Table~\ref{tab:bradap-friedman}.
According to this test, the most competitive methods are those which take into account information about nearer nodes (authorities and voters are closer than 100~m). This is because the closer surroundings change less frequently, and therefore, there is no need to continuously change the beacon rate.

The Swarm FREDY configuration $d_1$=0~m and $d_2$=50~m is significantly the most stable one. 
The least competitive results are those obtained by Swarm FREDY (200,250),
which is where it considers the highest number of nodes for the computations. 
The Wilcoxon statistical test applied to the two last ranked methods (Swarm FREDY (200,250) and Swarm DIFRA) reveals that there are no significant differences between them. 
Thus, it is important to select a suitable Swarm FREDY configuration to provide stable broadcasting.  

The Swarm FREDY variations studied provide more stable beacon rates (QoS).
This allows CVS applications to have a predictable performance. Thus, service designers are able to better adjust the applications to the real exchange of information between VANET nodes.

\subsection{Results Review}
\label{sec:results-review}

According to the experimental evaluation performed in this study, the answer to \textbf{RQ2} is yes, because
the proposed Swarm FREDY provides an efficient congestion control and QoS by using light cost computations.
This method improves
$i)$ the amount of data shared by the nodes (increases beacon rates),
$ii)$ the channel usage, and
$iii)$ the stability 
with regard to Swarm DIFRA.

Swarm DIFRA~\cite{Toutouh2016} provides competitive fair results because the nodes tend to broadcast beacons with lower and similar beacon rates. 
However, it is clearly the least competitive method for the other metrics analyzed. 
In answer to \textbf{RQ3}, the proposed stochastic method (Swarm FREDY) improved the performance of a deterministic congestion method based on similar fundamentals (Swarm DIFRA). 

\section{Conclusions and Future Work}
\label{sec:conclusions}

This article has studied the congestion control in VANET broadcasting.
More specifically, we have focused on the beaconing used by CVS applications, the most promising ones for road traffic safety and efficiency.
We have defined and tackled the FBR optimization problem, using automatic solvers.
%

We have devised Swarm FREDY, a swarm intelligence based family of algorithms which utilize light computations to address such an optimization problem dynamically.
As a swarm method, Swarm FREDY computations are based on combining self-monitored information (self-experience) and data received from the neighborhood (experience of the neighbors). 
Then, it allows an emergent behavior that leads the whole system to reliable, efficient, and useful communications for an updated system without a central authority. 

One of the main contributions of Swarm FREDY is the use of the stochastic distance discrimination procedure based on two parameters ($d_1$ and $d_2$).
This process classifies neighboring nodes into three different categories to get better and more simple information about the current network status, and thereby obtaining more accurate beacon rates.

The proposed algorithm family has been compared with the Swarm DIFRA method, which had demonstrated a competitive performance in comparison with other state-of-the-art congestion control methods in previous studies.
The experimental analysis confirms that significant improvements in congestion control are obtained when using Swarm FREDY, when compared with Swarm DIFRA.
When the VANET nodes use Swarm FREDY, they are able to communicate at higher beacon rates, enabling CVS applications to share information with high resolution.
In addition, the channel usage is maximized without exceeding its effective capacity, and therefore, the throughput is maximal. 
Finally, Swarm FREDY has demonstrated higher robustness than Swarm DIFRA, because the beacon rates computed by the first method vary less frequently than those computed by Swarm DIFRA.

There is some room for improvement in Swarm FREDY, as the results on balance, in experimental results versus DIFRA. 
The main lines of future research are three:
$i)$ evaluating the proposed congestion control method by using realistic urban scenarios aiming to confirm their competitive performance;
$ii)$ applying optimization strategies to compute the most promising (optimal) values for the configuration parameters of Swarm FREDY;
and
$iii)$ using the Swarm FREDY algorithms devised here as the starting point towards developing new distributed broadcasting methods that use modern computational intelligence strategies (e.g., Neural Networks).

\section*{References}

\bibliography{congestion-control-refs}

\end{document}